\documentclass[journal]{IEEEtran}
\ifCLASSINFOpdf
  \usepackage[pdftex]{graphicx}
\else
\fi
\usepackage{amsmath}
\usepackage{algorithmic}
\usepackage{array}
\ifCLASSOPTIONcompsoc
  \usepackage[caption=false,font=normalsize,labelfont=sf,textfont=sf]{subfig}
\else
  \usepackage[caption=false,font=footnotesize]{subfig}
\fi
\usepackage{url}
\begin{document}
%
% paper title
% Titles are generally capitalized except for words such as a, an, and, as,
% at, but, by, for, in, nor, of, on, or, the, to and up, which are usually
% not capitalized unless they are the first or last word of the title.
% Linebreaks \\ can be used within to get better formatting as desired.
% Do not put math or special symbols in the title.
\title{Continuously Constructive Deep Neural Networks}
%
%
% author names and IEEE memberships
% note positions of commas and nonbreaking spaces ( ~ ) LaTeX will not break
% a structure at a ~ so this keeps an author's name from being broken across
% two lines.
% use \thanks{} to gain access to the first footnote area
% a separate \thanks must be used for each paragraph as LaTeX2e's \thanks
% was not built to handle multiple paragraphs
%

\author{Ozan~\.Irsoy,
        Ethem~Alpayd{\i}n~\IEEEmembership{}% <-this % stops a space
\thanks{O. \.Irsoy is with Bloomberg LP, New York, NY 10022, e-mail: oirsoy@bloomberg.net.}% <-this % stops a space
\thanks{E. Alpayd{\i}n is with the Department of Computer Engineering, Bo{\u g}azi{\c c}i University, Istanbul 34342 Turkey, e-mail: alpaydin@boun.edu.tr.}% <-this % stops a space
%\thanks{Manuscript received April 19, 2005; revised August 26, 2015.}
}

\maketitle

% As a general rule, do not put math, special symbols or citations
% in the abstract or keywords.
\begin{abstract}
Traditionally, deep learning algorithms update the network weights whereas the network architecture is chosen manually, using a process of trial and error. In this work, we propose two novel approaches that automatically update the network structure while also learning its weights. The novelty of our approach lies in our parameterization where the depth, or additional complexity, is encapsulated continuously in the parameter space through control parameters that add additional complexity. We propose two methods: In tunnel networks, this selection is done at the level of a hidden unit, and in budding perceptrons, this is done at the level of a network layer; updating this control parameter introduces either another hidden unit or another hidden layer. We show the effectiveness of our methods on the synthetic two-spirals data and on two real data sets of MNIST and MIRFLICKR, where we see that our proposed methods, with the same set of hyperparameters, can correctly adjust the network complexity to the task complexity.\end{abstract}

% Note that keywords are not normally used for peerreview papers.

% For peer review papers, you can put extra information on the cover
% page as needed:
% \ifCLASSOPTIONpeerreview
% \begin{center} \bfseries EDICS Category: 3-BBND \end{center}
% \fi
%
% For peerreview papers, this IEEEtran command inserts a page break and
% creates the second title. It will be ignored for other modes.
\IEEEpeerreviewmaketitle

\section{Introduction}
% The very first letter is a 2 line initial drop letter followed
% by the rest of the first word in caps.
% 
% form to use if the first word consists of a single letter:
% \IEEEPARstart{A}{demo} file is ....
% 
% form to use if you need the single drop letter followed by
% normal text (unknown if ever used by the IEEE):
% \IEEEPARstart{A}{}demo file is ....
% 
% Some journals put the first two words in caps:
% \IEEEPARstart{T}{his demo} file is ....
% 
% Here we have the typical use of a "T" for an initial drop letter
% and "HIS" in caps to complete the first word.
\IEEEPARstart{D}{eep} learning involves multiple layers of nonlinear information processing~\cite{Bengio-2009}. This allows learning architectures that implement functions as repeated compositions of simpler functions, thereby learning layers of abstraction with better generalization and representation capacity. Recent advances in efficient training of deep neural networks using larger data sets, GPUs, or better optimization techniques enabled their application to many problems, ranging from speech and vision to natural language processing.

Though deep learning is helpful, having too many layers may be problematic: First, when there are more layers/units/weights, space and computational complexity is higher; second, when there are more free parameters, there is a higher risk of overfitting, which needs to be combated using regularization methods such as weight sharing, weight decay, dropout, and so on. Third, when the network is deep, there is the problem of vanishing/exploding gradients when the error is backpropagated over many layers~\cite{bengioVanishingGradient}, and one relatively simple mechanism is to introduce gating mechanisms that allow passing signals with minor modification, as in LSTM or GRU for sequence learning~\cite{lstm,chung2014empirical}.

Starting from 1990s, there have been many approaches to optimize the network architecture, ranging from the early incremental methods of adding hidden units one by one~\cite{ash1989dynamic, frean1990upstart}, or starting from a large network and pruning it~\cite{reed1993pruning}, to more complex recent approaches, such as evolutionary algorithms~\cite{jozefowicz2015empirical} or reinforcement learning~\cite{zoph2016neural}, as well as boosting-style methods~\cite{cortes2016adanet}.

Our work here has the similar goal of learning the network architecture from data. The main difference in our work is that instead of searching over a discrete space of all architectures, we parameterize our models in such a way that the notion of complexity, or depth, itself is continuous, making the model end-to-end differentiable, and allowing gradient-descent to search over the architectures in addition to their parameters.

We propose two methods for continuously constructing deep neural networks: In tunnel networks, inspired from highway networks~\cite{highway}, associated with each hidden unit is a continuous parameter and if this parameter is not active, the unit just copies its input to its output bypassing the nonlinearity. We start with a network with many layers with all its hidden layers inactive; during gradient-descent, the parameter may change and the corresponding unit becomes active and this implies adding a new layer of nonlinearity effectively increasing the depth of the network. 

In our second method of budding perceptrons, inspired from budding trees~\cite{budding}, there is a parameter associated with each layer indicating whether further nonlinear processing is needed. Initially we start with a single layer and during learning with gradient-descent, when needed, this parameter may become active which causes the creation of another full layer, effectively increasing the depth of the network.

We start by a survey of algorithms that adapt network structure during learning in Section~\ref{sec:related}, after which we explain our two methods in detail in Section~\ref{sec:methods}. Our experiments are discussed in Section~\ref{sec:experiments} where we first show our didactic results on the synthetic, two-dimensional two-spirals data, and then on the larger, real-world data sets of MNIST and MIRFLICKR. Our results indicate that both methods can grow networks automatically to learn these problems; with the same set of hyperparameters, the complexity of the learned network matches to the complexity of the underlying task. For simple problems, small networks are constructed and the depth increases as the complexity of the task it faces increases. We conclude and discuss future work in Section~\ref{sec:conclusions}.

\section{Related Work}
\label{sec:related}

The optimization of the network structure is typically done manually using trial and error. A straightforward way of searching for the architecture is to treat the number of layers and units as hyperparameters and optimize over these hyperparameters \cite{bergstra2011algorithms, bergstra2012random, snoek2012practical, snoek2015scalable}. 

Early work on searching for the optimal network focused on incremental methods. Most such methods fix the depth to one and so the problem reduces to a search over the number of hidden units~\cite{ash1989dynamic, frean1990upstart, friedman1981projection, setiono2001feedforward}. Among these, there are those that add hidden units one by one; such as dynamic node creation~\cite{ash1989dynamic}, upstart algorithm~\cite{frean1990upstart}, and feedforward neural network construction using cross-validation~\cite{setiono2001feedforward}. A related method is cascade-correlation, where a new hidden unit that has the rest of the network as input, is trained while the rest of the network weights are fixed~\cite{fahlman1990cascade, lahnajarvi2002evaluation, lehtokangas1999fast, lehtokangas2000modified}. This process technically increases the depth of the network for each addition of a unit. A recent related work applies a similar methodology to convolutional networks where a pre-trained network is widened or deepened on a new task to achieve better knowledge transfer~\cite{wang2017growing}.

Pruning methods start with a large network and remove units/connections if they are deemed unnecessary~\cite{castellano1997iterative, lauret2006node, ponnapalli1999formal, reed1993pruning}. After training a network, weights may be set to zero if the change in generalization error is small. Weight decay is equivalent to L2 regularization and other regularization and Bayesian techniques are also used for model selection in neural networks \cite{ma2004gradient, vila2000bayesian, xu2003byy}.

There are also hybrid methods that allow both addition and deletion of hidden units and layers \cite{nabhan1994toward}. Many hybrid methods are proposed in the context of radial-basis function networks where hidden units have local responses~\cite{oliveira2005improving, paetz2004reducing, huang2005generalized}. Finding the right network can be defined as a search in the space of possible networks with operators that add/remove units/layers and different search methods can be used here \cite{aran2009incremental}. One can also employ evolutionary search for a wider search~\cite{leung2003tuning, li2006system, macleod2001incremental, zhu2005evolutionary, jozefowicz2015empirical}.

Another direction is to use reinforcement learning in which a controller network that generates candidate neural network descriptions is trained to optimize the expected accuracy~\cite{zoph2016neural}. More recently, a boosting-style method named AdaNet is used to incrementally grow architectures while learning the parameters~\cite{cortes2016adanet}; AdaNet also provides data-dependent generalization guarantees of their method.

\section{Proposed Methods}
\label{sec:methods}
\subsection{Tunnel Networks}

Our first approach involves defining a parameter, $g\in [0,1]$, for each hidden unit that determines how much it acts as a nonlinear unit versus how much it just copies its input to its output without performing any transformation on it:
\begin{align}
y(x) = g \cdot  \sigma(w^Tx + b) + (1-g) \cdot x
\label{eq:tunnel}
\end{align}

If $g=0$, the unit just copies its input to its output; if $g=1$, the output is given by the output of the nonlinear $\sigma(w^Tx + b)$ with its internal parameters of $w$ and $b$, namely the weights and biases. We have a layer of such units that takes an (elementwise) convex combination of a perceptron layer and the identity mapping. 

We can consider $g$ as a gating parameter that chooses between identity and the nonlinear transformation. We can also interpret it as a complexity control parameter that chooses between the simpler identity and the more complex nonlinear hidden unit. As long as it is 0, we have a simple model, but if $g$ gets larger during learning, the nonlinearities start to kick in, the network becomes more complex, and one can interpret this as growing a neural network constructively. 

To favor simplicity, we use an objective function that penalizes positive values for $g$:
\begin{align}
    J(X) = \sum_{x\in X} E(x) + \lambda \sum_{l} g_l
\end{align}
where $x$ is an instance from the dataset $X$, $J(\cdot)$ is the overall objective function to be minimized, $E(\cdot)$ is the error (loss) function over an instance (e.g. cross-entropy), $l$ is the index of each unit in all layers and $\lambda \geq 0$ is the trade-off parameter adjusted using cross-validation. During stochastic gradient-descent, we update all $w_l$ and $b_l$ in the network and also the $g_l$ values. During learning, just as it is possible that $g_l$ moves away from 0 and adds a nonlinear hidden unit, it is also possible that because of regularization, it moves back to 0 effectively pruning it back.

Tunnel networks are inspired from highway networks~\cite{highway} where $g$ is a function of $x$ and as such works as an input-dependent gating model:
\begin{align}
    g(x)= \sigma_g(w_g^Tx + b_g) 
\end{align}

We can consider our tunnel network as a special case of highway networks where the gating is constant. In highway networks, a negative initial bias for $b_g$ is used, such as $-2$ or $-4$ \cite{highway}, which results in starting values for $g(x)$ that are close to 0, and hence each individual layer starts close to the identity. As $g(x)$ values move away from 0 when $w_g$ and $b_g$ are updated, the network response becomes nonlinear. A highway network where $g(x)$ values are regularized to be close to 0 (by having negative $b_g$ and very small $w_g$) would work similarly to our tunnel network.

Another related architecture is the deep residual network~\cite{he2016deep} where there is no explicit gating and the transformation and the input copy are summed directly. That is, in Eq.~\ref{eq:tunnel}, both $g$ and $(1-g)$ terms are set to 1 (for a residual block of a single layer perceptron).
Deep residual networks have won several competitions including ImageNet and COCO challenges~\cite{he2016deep}, indicating the strength of methods that can adjust network complexity to that of data during training.

\subsection{Budding Perceptrons}

Our second approach is inspired by the budding tree, which we proposed before, and it is a tree where complexity is softly parameterized~\cite{budding}. Unlike a regular tree where a node is either a decision node or a leaf node, in a budding tree each node $m$ has a leafness parameter $\gamma_m\in [0,1]$ and is both a leaf node with weight $\gamma_m$ and an internal node with weight $1-\gamma_m$. In the beginning, the root is a leaf (its $\gamma$ is equal to 1), but when it is updated (using gradient-descent) to get smaller, its two children are created, effectively growing the tree. The error function has a regularization term to penalize $\gamma_m$ that are smaller than 1. 

In the budding perceptron we propose here, the approach is the same except we have perceptron layers in a network instead of nodes of a tree. Every perceptron layer has a complexity parameter $\gamma_m$ which is initially 1 and if $\gamma_m$ gets smaller, another full perceptron layer is created next to it.

More formally, a budding perceptron network uses a tree structure to implement the composition of individual perceptron layers to make up a deep network. Its recursive definition is as follows:
\begin{align}
    y_m(x) = (1-\gamma_m) \cdot y_{mr}(y_{ml}(x)) + \gamma_m \cdot \sigma (w_m x + b_m) 
\end{align}

\noindent where $ml$ and $mr$ are the left and right children of $m$. $\gamma \in [0,1]$ is the complexity parameter and $\sigma$ is the nonlinearity function used in the hidden layers. If $\gamma_m$ is 1, we only have the current perceptron layer, but for $\gamma_m<1$, its children layers are also created and every parent becomes a convex combination of a perceptron and the composition of two similar functions implemented by its children.\footnote{We apologize for the possible confusion: In the tunnel network, $g_l=0$ is the simpler alternative and $g_l=1$ is the more complex, whereas in the budding perceptron, $\gamma_m=1$ is the simpler alternative, i.e., a perceptron, and $\gamma_m=0$ is the complex alternative of two children perceptrons. This is because we wanted to stick to the definitions in the original models of highway networks and budding trees from which we are inspired.}

Observe that the parameterization above defines an infinite complete binary tree since every node in the tree has two children. Therefore the parameter space contains feedforward networks of all depths. In practice, since all $\gamma_m$ values start from 1 and during training only finitely many of them will assume values that are smaller, the recursion is guaranteed to end when a node with $\gamma = 1$ is encountered and its children will not be evaluated.

We add a regularizer term to prefer smaller networks :
\begin{align}
    J(X) = \sum_{x\in X} E(x) + \lambda \sum_{m \in T} (1-\gamma_m)
\end{align}
where $T$ denotes the set of all nodes in the tree.

During gradient-descent, we update all the perceptron weights, $w_m$ and $b_m$, and also $\gamma_m$. Initially, a single root node is equivalent to a single layer of perceptron with its $\gamma$ equal to 1. During training, when its $\gamma$ is updated to get smaller, two perceptron layers are created as children (with their own $\gamma_m$ initially 1) making up a deeper network. Then in turn those two layers can spawn new layers, effectively adding more and more layers. It is important to note that during gradient-descent, $\gamma_m$ values move away from 1 adding new layers and then as the network weights converge, $\gamma_m$ may move back to 1 effectively eliminating previously added layers.

Note that with the above definition, there is no information passed on from parent weights (e.g. $w_m$) to weights of the children (e.g. $w_{ml}$). This means that when a leaf is split into two nodes, $w_{ml}$ and $w_{mr}$ need to be learned from scratch. This might slow down convergence considerably and impact learning dynamics negatively. To help with information propagation when splitting, we share (tie) the weights of one of the children with the parent.

\subsection{Measuring Network Complexity}
\label{sec-complexity}

Since we have notions of complexity that are soft, discrete measures such as depth or the number of hidden units are not directly applicable. For instance, a ten-layer tunnel network that has all $g$ values equal to 0 implements an identity function, whereas a same size tunnel network with $g$ values equal to 1 is a multilayer perceptron with depth 10. 

To this end, we propose soft criteria for measuring the complexity of tunnel networks, highway networks and budding perceptrons we use in our experiments. For tunnel networks, it is simply the layerwise (or total) sum of all $g$ values:
\begin{align}
    s_l = \sum_k^K g_{lk} & \text{ for layer $l\in L$}\mbox{\ and\ }
    s = \sum_{l\in L} s_l
\end{align}
where $l$ indexes the layer out of the set $L$ of all layers, and $k$ indexes a hidden unit out of the $K$ hidden units of layer $l$.

For highway networks, $g_{lk}(x)$ replaces the constant $g_{lk}$ and since it is a function of the input, we compute it as the average over all input instances (in the validation set):
\begin{align}
    s_l(X) &= \sum_k^K \dfrac{1}{|X|}\sum_{x\in X} g_{lk}(x) & \text{for layer $l\in L$}\\
    s(X) &= \sum_{l \in L} s_l(X)
\end{align}

For budding perceptrons, in addition to the tree size (node counts), we define a \emph{soft tree size} that incorporates the leafness parameters $\gamma_m$ as well:
\begin{align}
    s_m &= 1 + (1-\gamma_m)(s_{ml} + s_{mr}) & \text{for node $m$}\\
    s &= s_\text{root}
\end{align}

This soft size gives us a notion of effective depth of a budding perceptron network since every node itself acts as a single
(nonlinear) perceptron layer.

\begin{figure}[!htbp]
\centering
    \includegraphics[width=0.75\columnwidth]{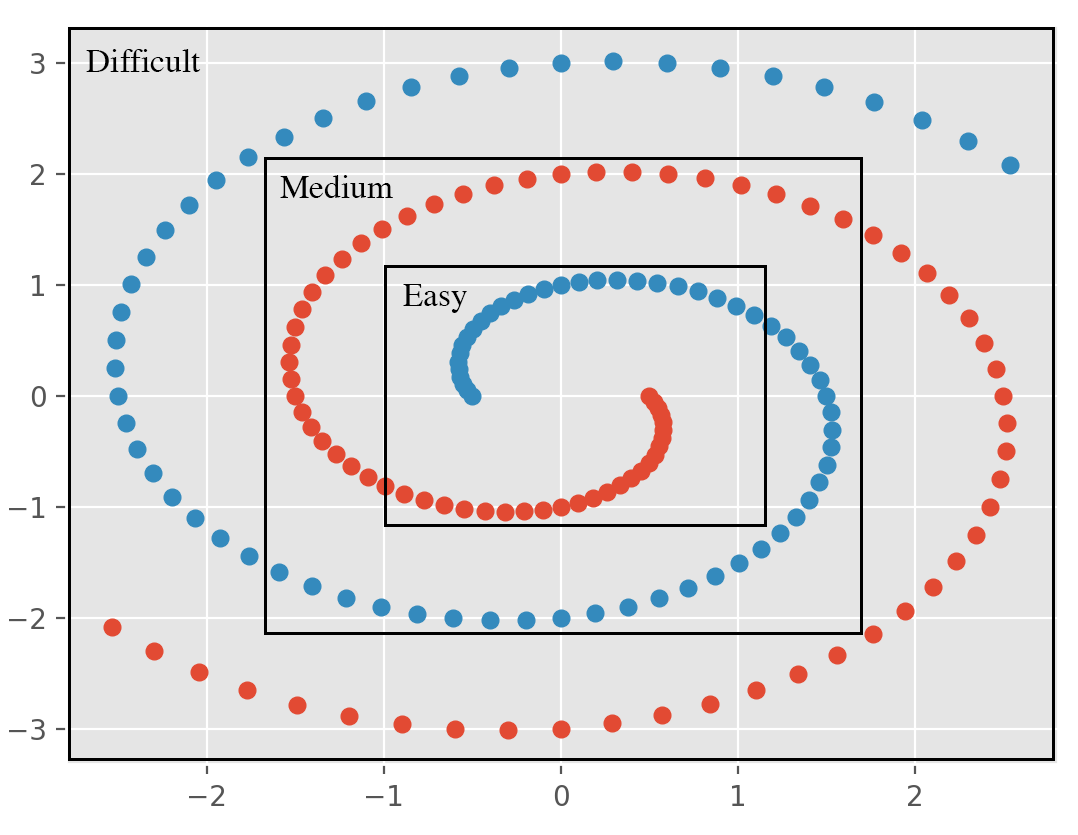}
\caption{Three different variants of the two spirals data set in increasing complexity. Different colors show different class labels.}
\label{fig-twospi}
\end{figure}

\begin{figure*}[!htbp]
\centering
\subfloat[Easy]{
\includegraphics[width=0.3\textwidth]{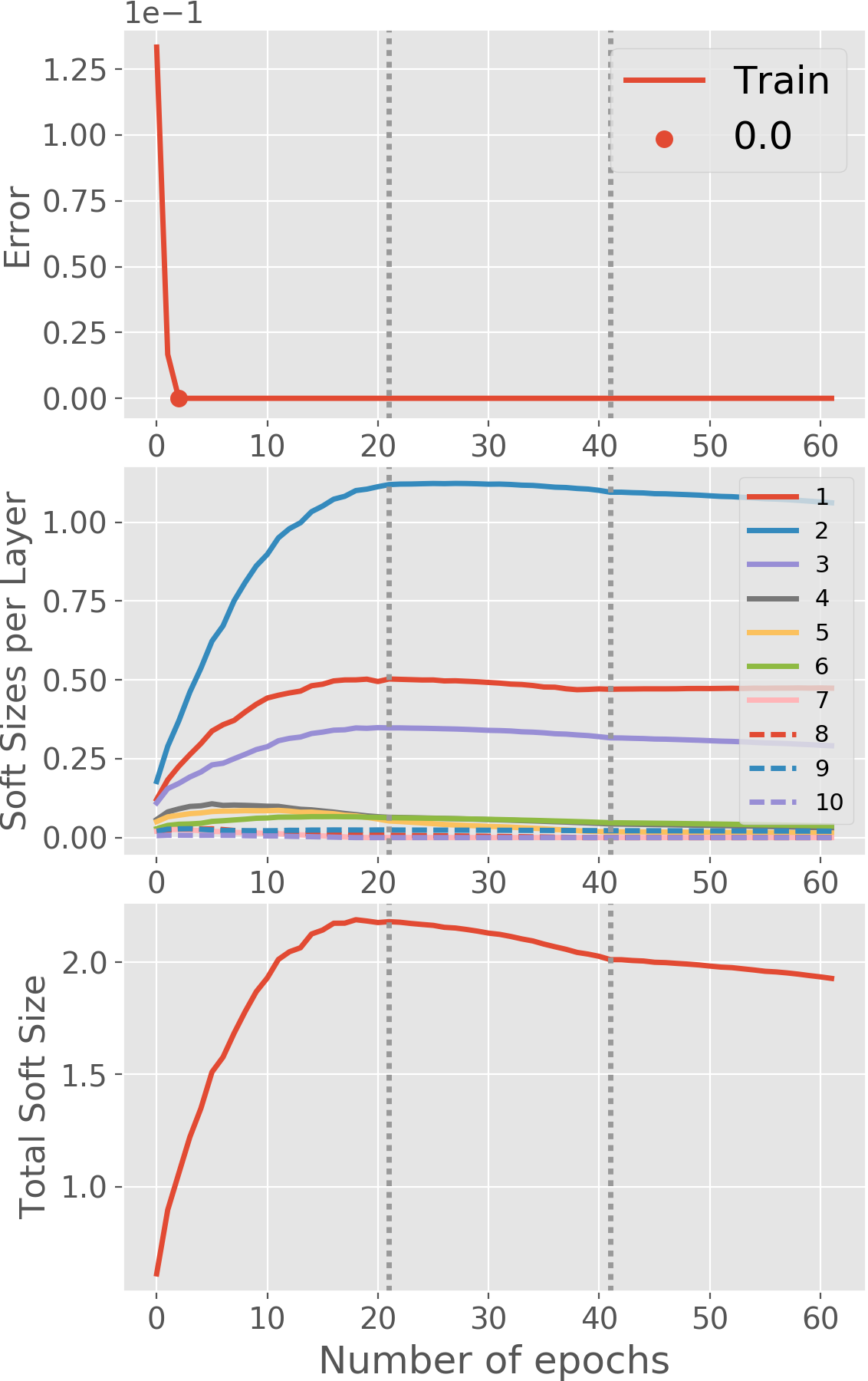}
}
\subfloat[Medium]{
\includegraphics[width=0.3\textwidth]{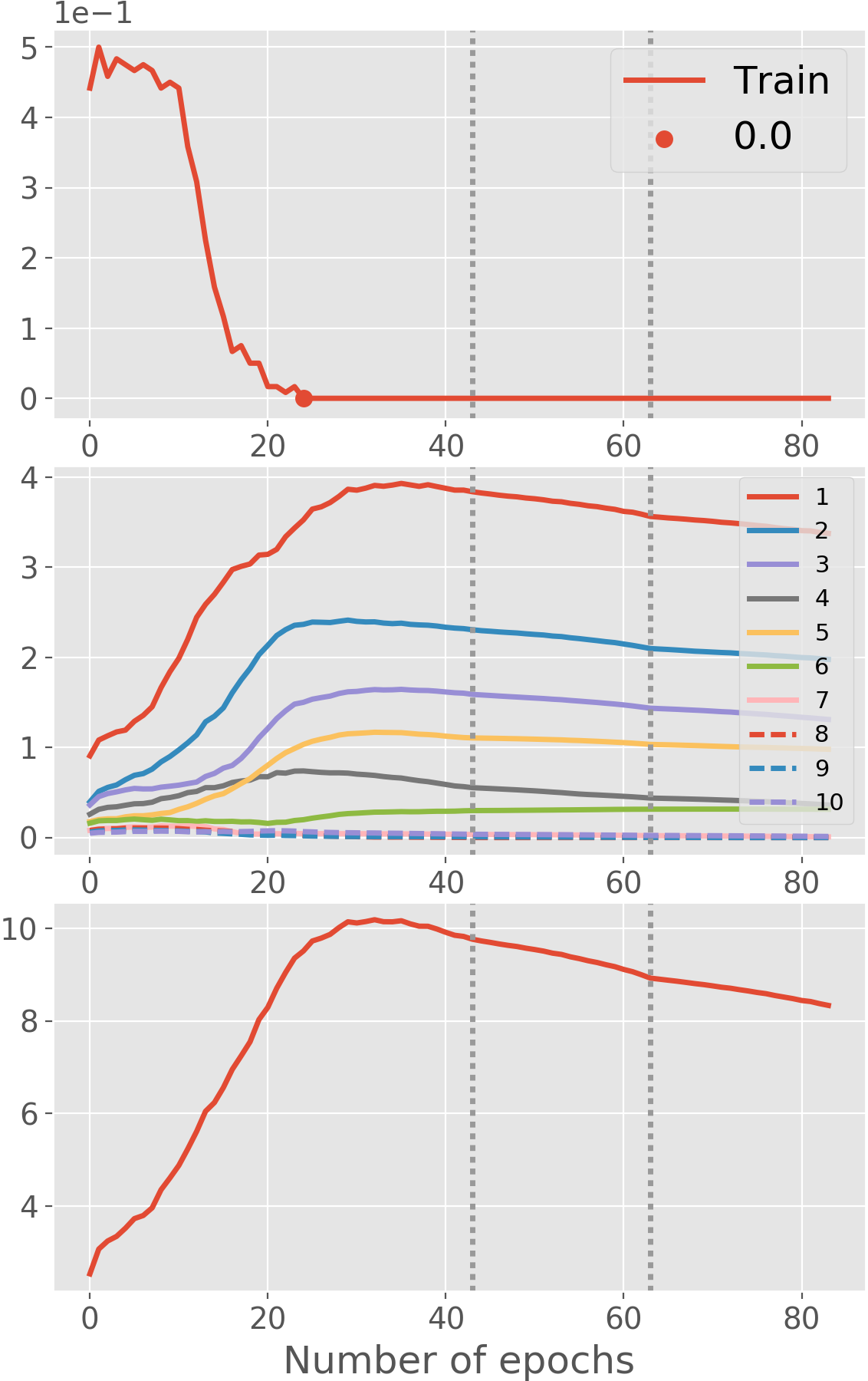}
}
\subfloat[Difficult]{
\includegraphics[width=0.3\textwidth]{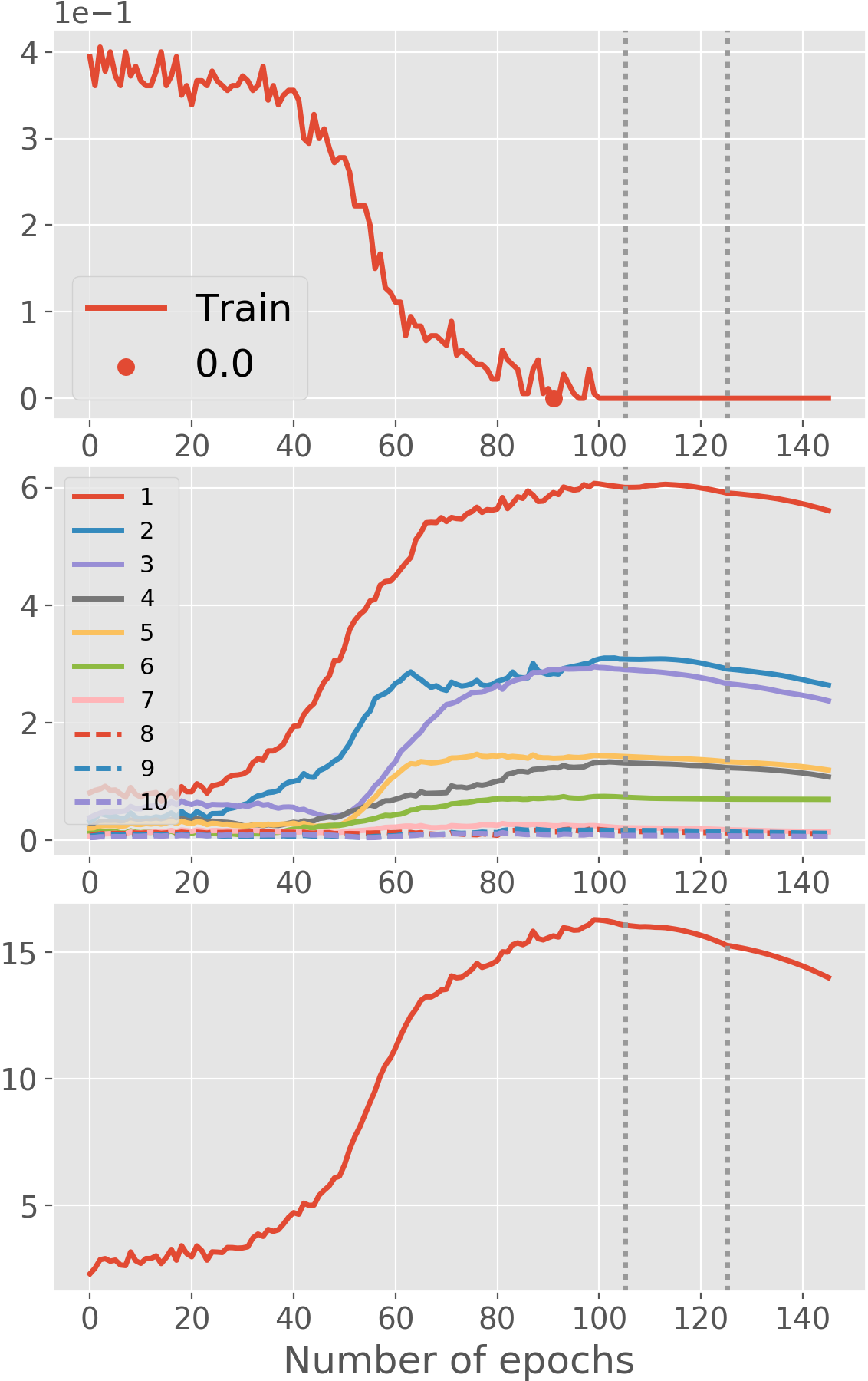}
}
\caption{Results of tunnel networks on different variants of two-spirals.We show error rate (top), total $g$ values
for each layer individually (middle) and for all layers (bottom). Throughout these figures, vertical bars show where learning rate shrinkage occurs. During training, complexity increases but may start decreasing back again. }
\label{fig-twospi-budmlp}
\end{figure*}

\begin{figure*}[!htbp]
\centering
\centering
\subfloat[Easy]{
\includegraphics[width=0.3\textwidth]{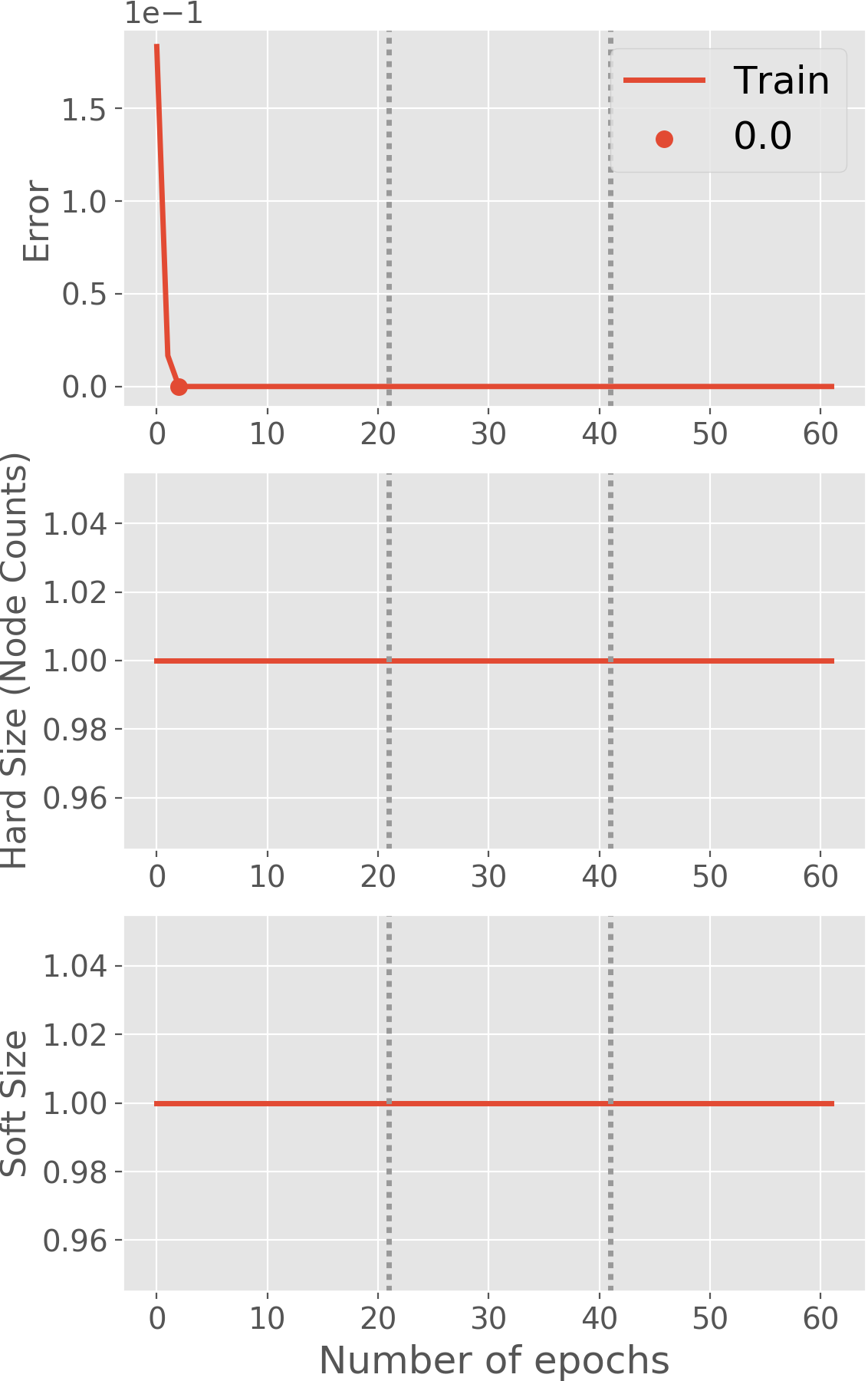}
}
\subfloat[Medium]{
\includegraphics[width=0.3\textwidth]{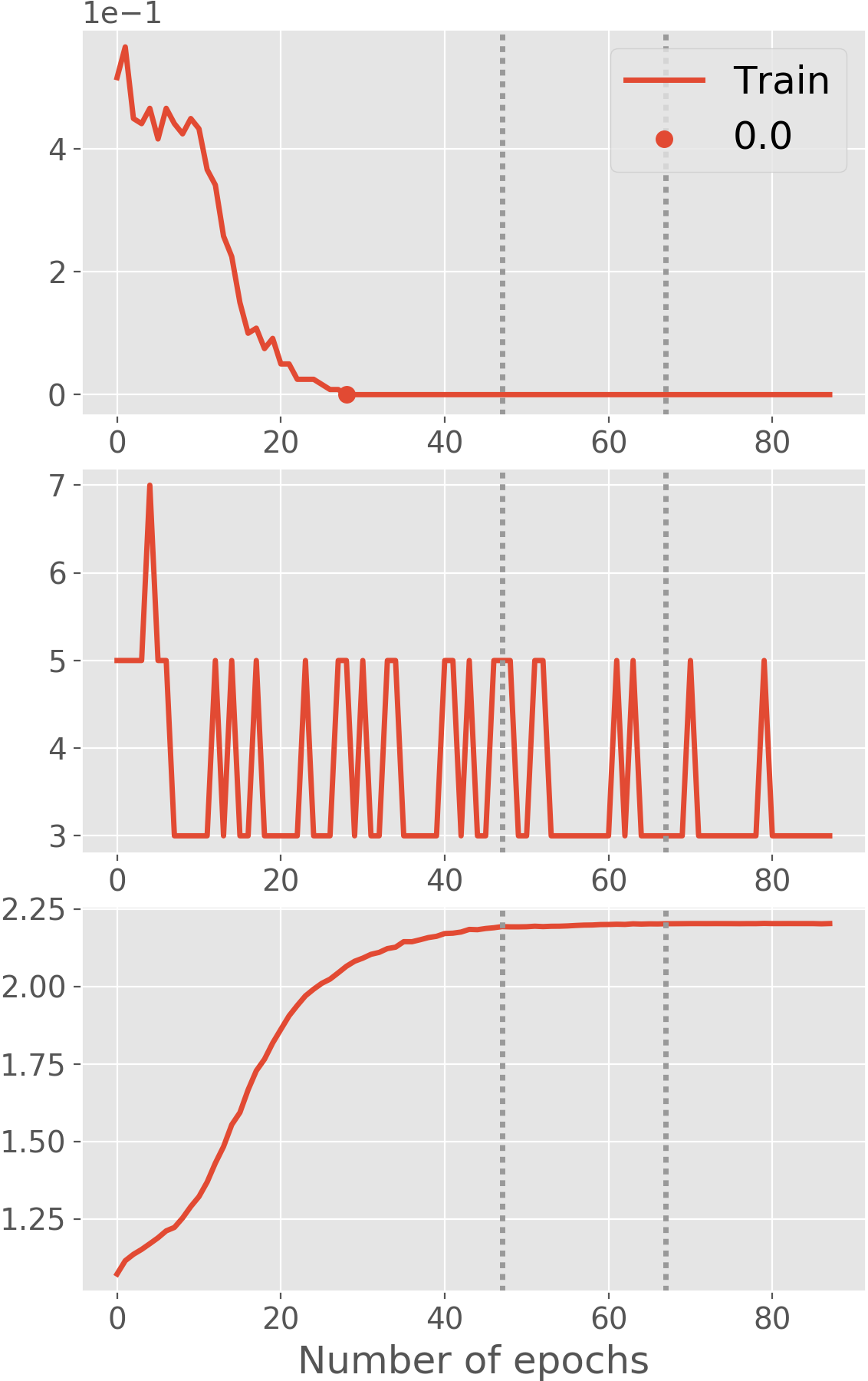}
}
\subfloat[Difficult]{
\includegraphics[width=0.3\textwidth]{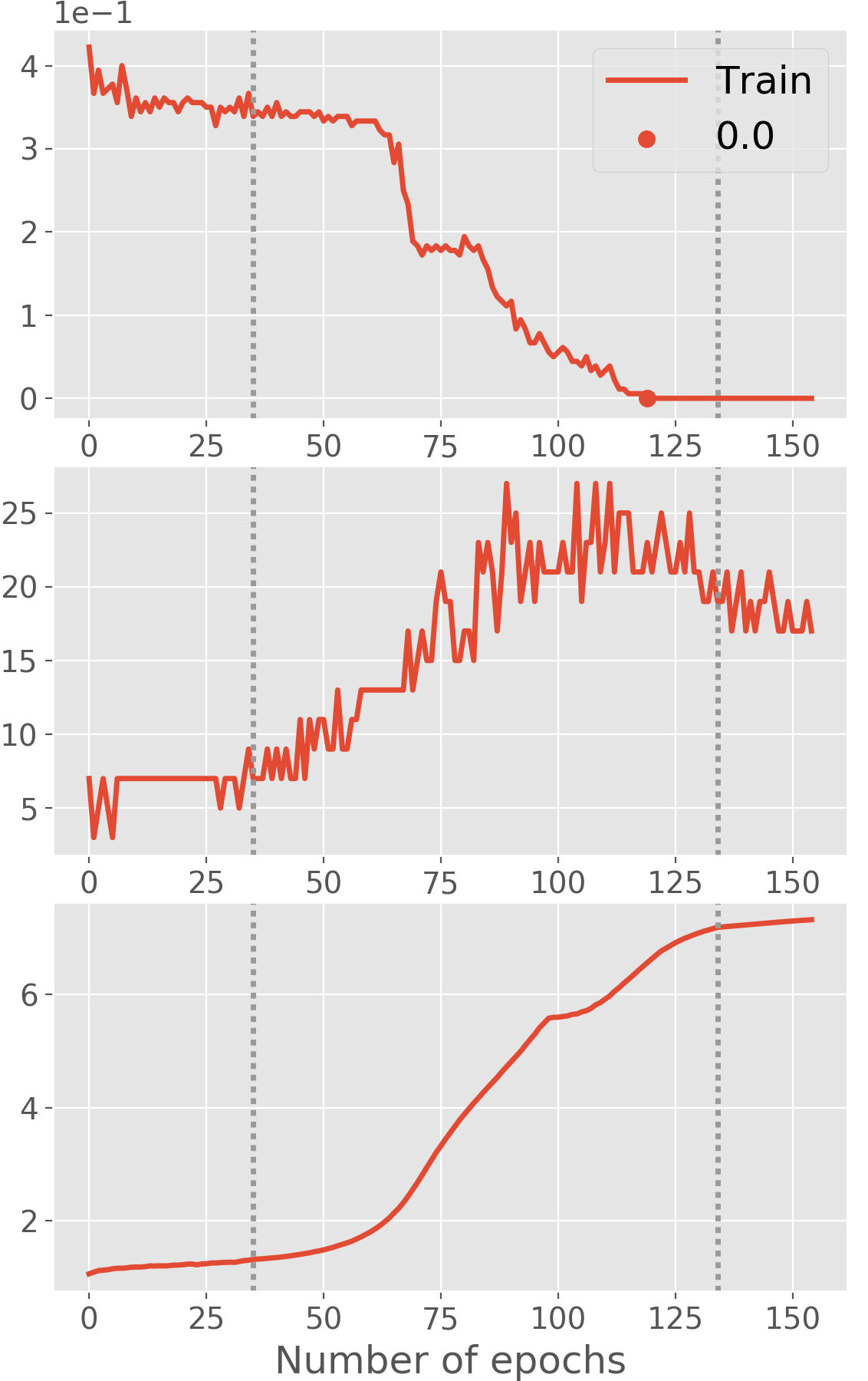}
}
\caption{Results of budding perceptrons on the different variants of two-spirals. We show error rate (top), hard tree sizes (middle), and
soft tree sizes (bottom).}
\label{fig-twospi-btn}
\end{figure*}

\begin{figure*}[!htbp]
\centering
\subfloat[Easy]{
\includegraphics[width=0.2\textwidth]{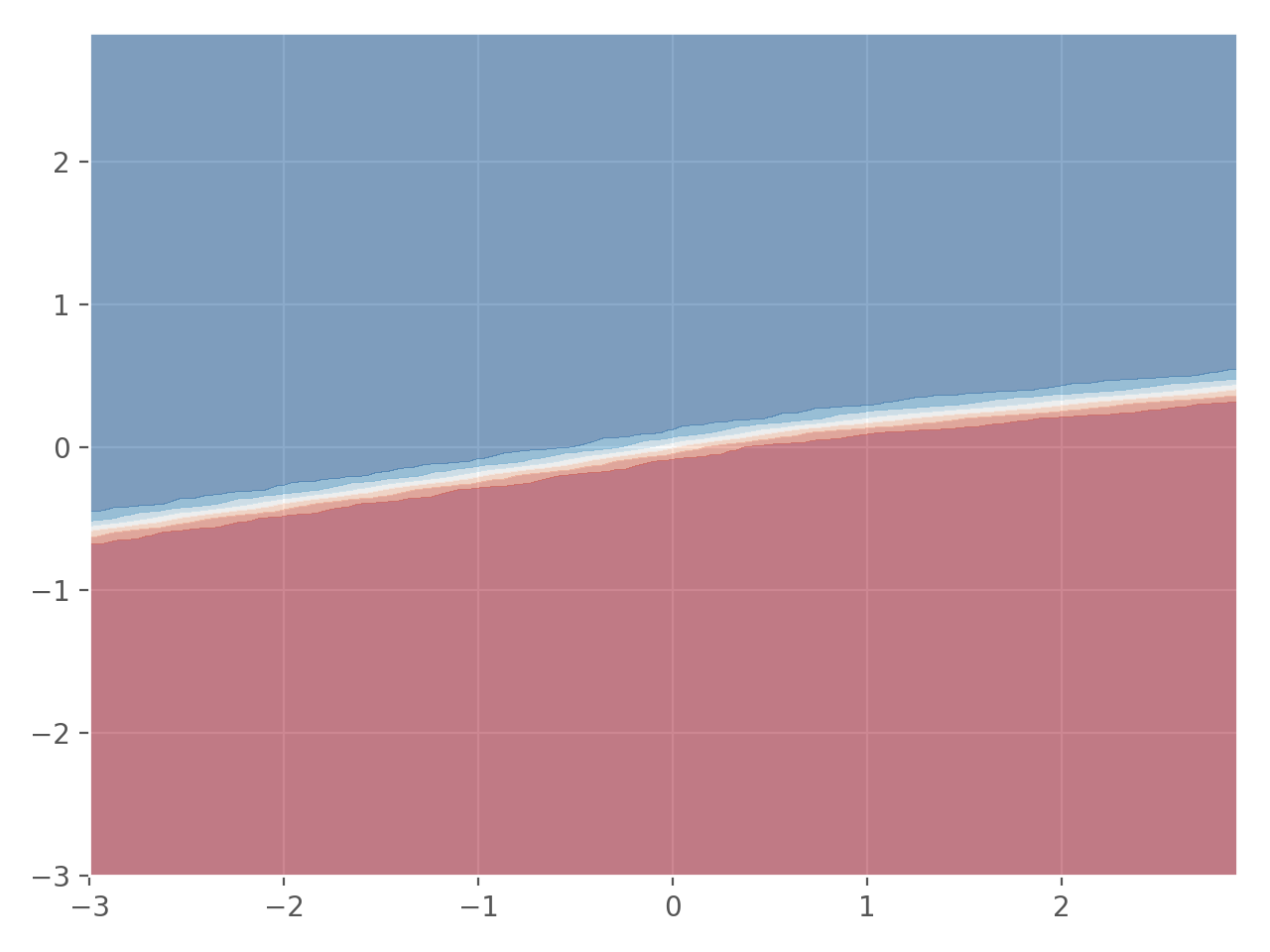}
}
\subfloat[Medium]{
\includegraphics[width=0.2\textwidth]{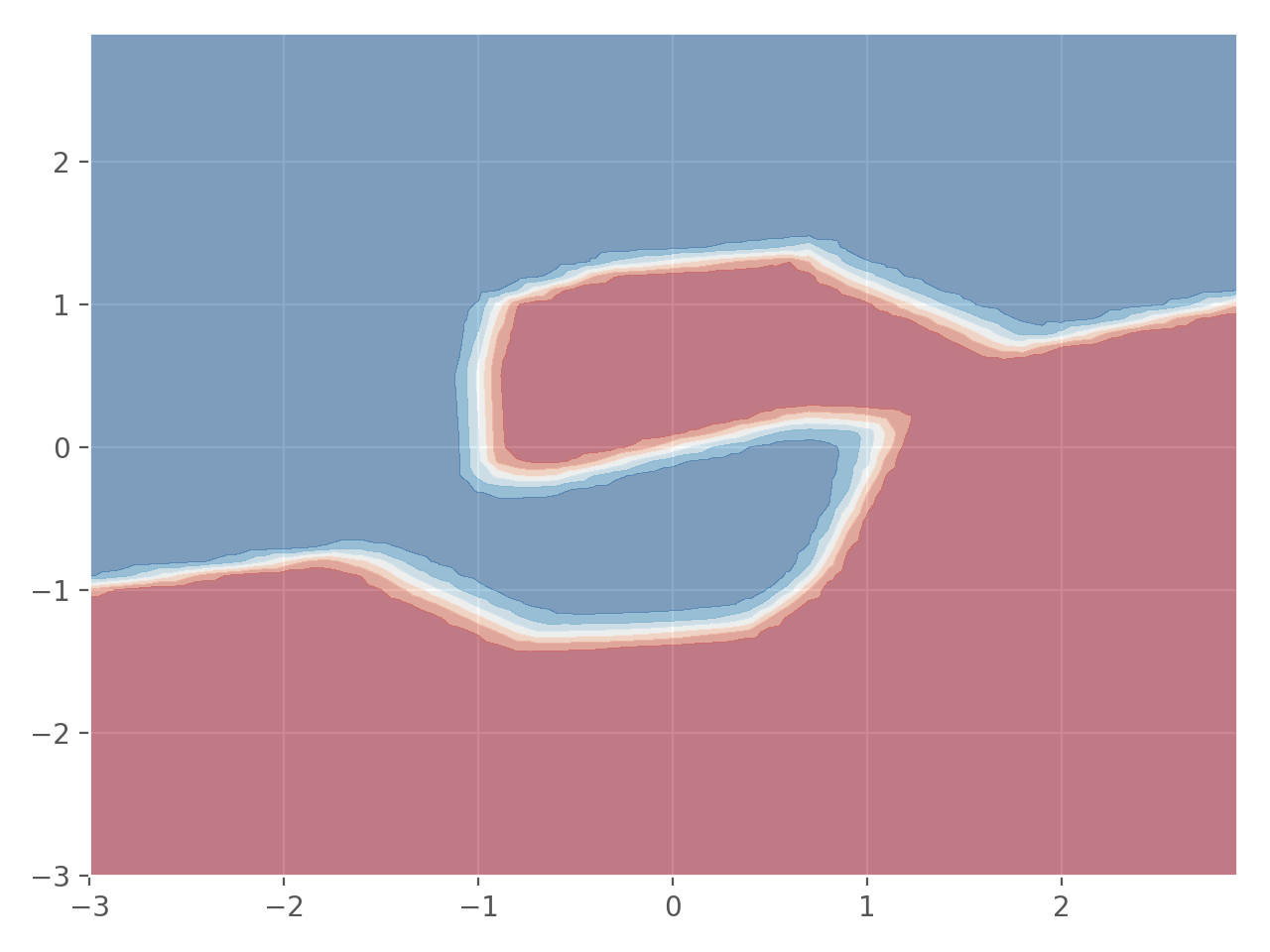}
}
\subfloat[Difficult]{
\includegraphics[width=0.2\textwidth]{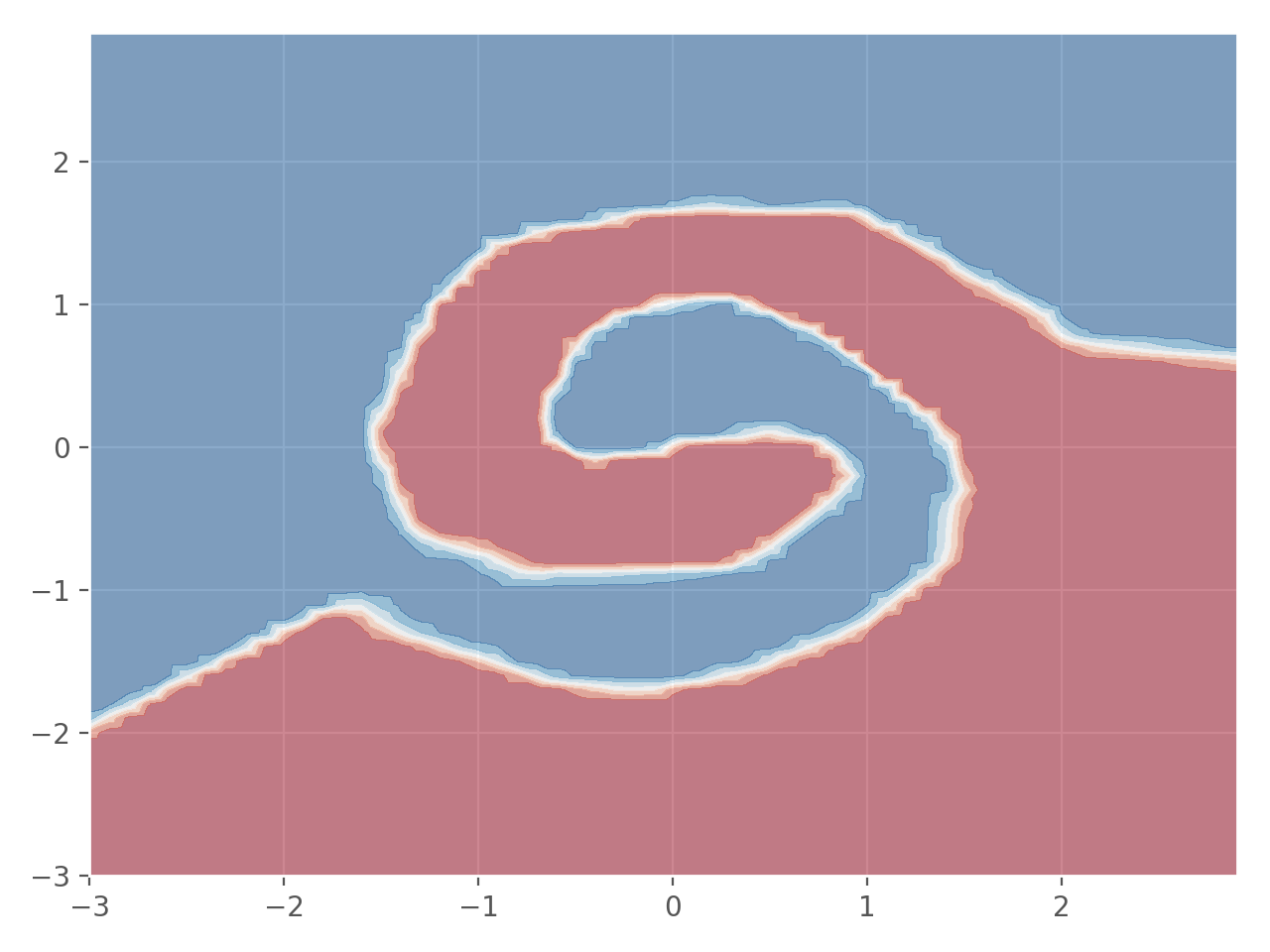}
}
\caption{Discriminants learned by budding perceptrons on the different variants of the two spirals data.}
\label{fig-twospi-btn-predict}
\end{figure*}

\section{Experiments}
\label{sec:experiments}

\subsection{Results on Two Spirals}

\textbf{Data.} 
We use the two spiral dataset, which is a two-class classification task \cite{lang1988learning}, for didactic experimentation. Our claim is that our constructive methods can adjust the complexity of the learned network to the complexity of the problem. To test this, we created three variants of this task, namely easy, medium and hard, as shown in Figure~\ref{fig-twospi}. Note that the easy task is linearly separable and the medium and hard variants need deeper networks.

\textbf{Training.} First off, it is important to stress that we use the same hyperparameters for different variants of the two spirals and we want to see if the constructive algorithm can find the right complexity just by changing the data.

We train all models with stochastic gradient-descent using the Adam update rule, which modifies learning rates and momentum rates for each parameter during training~\cite{kingma2014adam}. We use a binary cross-entropy objective function ($\sum_x \mathbf{1}(r(x) = 1) \log(y(x)) + \mathbf{1}(r(x) = 0)(1-\log(y(x))$ where $r(x) \in \{0, 1\}$ is the true label of $x$) on top of a sigmoid output layer. We do not employ mini-batching for the two spirals data but use the purely online setting. We schedule learning rate and determine the length of training as follows: Training continues until there is no improvement on the development set (which is the same as the training set for this synthetic dataset) for 20 epochs after which we use 0.3th of the original learning rate. Next time in which there is no improvement for 20 epochs we switch to 0.1th of the original learning rate. Finally, training stops when there is no further improvement for another 20 epochs.

We assign ten hidden units to each layer of tunnel and highway networks and budding perceptrons. For tunnel networks, we use a total (maximum) number of ten layers. We use a penalty of 0.001 for $\lambda$ in L1 regularizers. For learning rate, we pick the smallest one from the set \{0.0001, 0.0003, 0.001, 0.003\} that yield zero training error which results in 0.003 for tunnel networks and 0.001 for budding perceptrons. We employ a standard L2 regularization penalty of $10^{-5}$ on network weights which is typically useful for avoiding parameter explosions.

We tried two approaches for making learning rate dependent on depth. In the first case, we simply use the same learning rate on all layers. We found that this typically results in similar behavior for all layers (in terms of soft size growth during training) and layers are not distinguished. To better differentiate the layer behavior, we assigned decreasing learning rates to each layer, where each layer has 3/4th of the learning rate of the previous layer. This results in a more incremental behavior where lower layers grow more quickly and higher layers adapt slowly. Furthermore, as we will see in the experiments, the learning dynamics in this case encourages lower layers to be used early on, and if higher layers are not necessary, they are not used at all, allowing them to be actually pruned from the network after training.

All networks use rectifier activations ($\max\{0, x\}$) for their elementwise nonlinearities~\cite{glorot2011deep}. Since all of the models we evaluate require input and output dimensionality to be the same, all architectures have an additional linear projection layer at the very beginning. For this instance, since we have ten hidden units and the input space is two dimensional, we have a $2 \times 10$ matrix that linearly projects an input instance to a 10d space (which is also learned during backpropagation).

\textbf{Results.}
Our plots using tunnel networks are shown in Figure~\ref{fig-twospi-budmlp} and budding perceptrons are shown in Figure~\ref{fig-twospi-btn}. The horizontal axis corresponds to the number of iterations (measured in the number of epochs). 

In the top row of Figure~\ref{fig-twospi-budmlp}, we see the training error rates for easy, medium and difficult cases, respectively. In all cases, tunnel networks are able to converge to an error of zero. However, as the problem becomes more difficult, zero error is reached more slowly---it takes 2, 23, and 91 epochs to reach zero error for easy, medium and difficult cases, respectively.

In the middle row, we plot soft sizes for layers, each layer in a separate color. We see that early on, the constructive method introduce more layers than necessary but as learning proceeds those extra ones are pruned back. The peak is reached around where the models reach zero error, which is to be expected: Since cross-entropy and misclassification error are highly correlated, from then on, the objective improves mostly by reducing the regularization term. We believe that this is an interesting end result of defining complexity using continuous parameters. There are no distinct, separate steps of structure growing and pruning, nor structure learning versus parameter learning; the network structure is learned, by growing or pruning, coupled with the learning of the parameters. We also observe that, typically, the model prefers to use earlier layers more. This is an artifact of the way we set our learning rates in a decreasing fashion: In preliminary experiments where we set the learning rates equally, we observed that all layers are used in similar amounts.

Finally, the bottom row shows the total complexity. We see that as the problem becomes more difficult, the model grows more in terms of its soft size. This suggests that the models indeed grow networks of different complexities depending on the complexity of the problem itself, with the same set of hyperparameters. For instance, soft sizes of the tunnel network are around 2, 10, and 15 for the easy, medium, and difficult tasks, respectively.

In the top row of Figure~\ref{fig-twospi-btn}, we see the training errors of budding perceptrons on easy, medium and difficult tasks, respectively. Similar to tunnel networks, budding perceptrons can reach zero error and the time to reach there increases with difficulty (about 2, 30 and 125 epochs, respectively).

The middle and bottom rows show the hard and soft sizes of the budding perceptrons. Similar to tunnel networks, trees grow to different sizes depending on the complexity of the problem. Budding perceptron uses the smallest possible size of 1 node on the easiest task whereas it uses soft sizes of roughly $2.2$ and $7$ on the medium and hard variants.\footnote{Note that because of our parameterization, smallest depth we can have is 1 for budding perceptrons, even though the problem can be solved with a depth of 0. A parameterization that started with the identity mapping at the root node (instead of a nonlinear perceptron) would allow the model to assume 0 depth.}

We see the discriminants learned for the different variants using the budding perceptron in Figure~\ref{fig-twospi-btn-predict}. We observe that the decision boundary (implemented by the same constructive model with the same set of hyperparameters) becomes more complex as needed, as the problem becomes harder (and in the easiest case, it is linear as expected). 

Tunnel networks also find similar boundaries adapted to the task. We have also tested highway networks on the two spirals data set. We cannot include those figures here because of lack of space, but we find that the highway network proper requires almost twice as many units as tunnel networks; when a regularization term is added (so that highway units act as tunnel units), the number of units become comparable.

\begin{figure}[!htbp]
\centering
\subfloat[Two-class]{
\includegraphics[width=0.49\columnwidth]{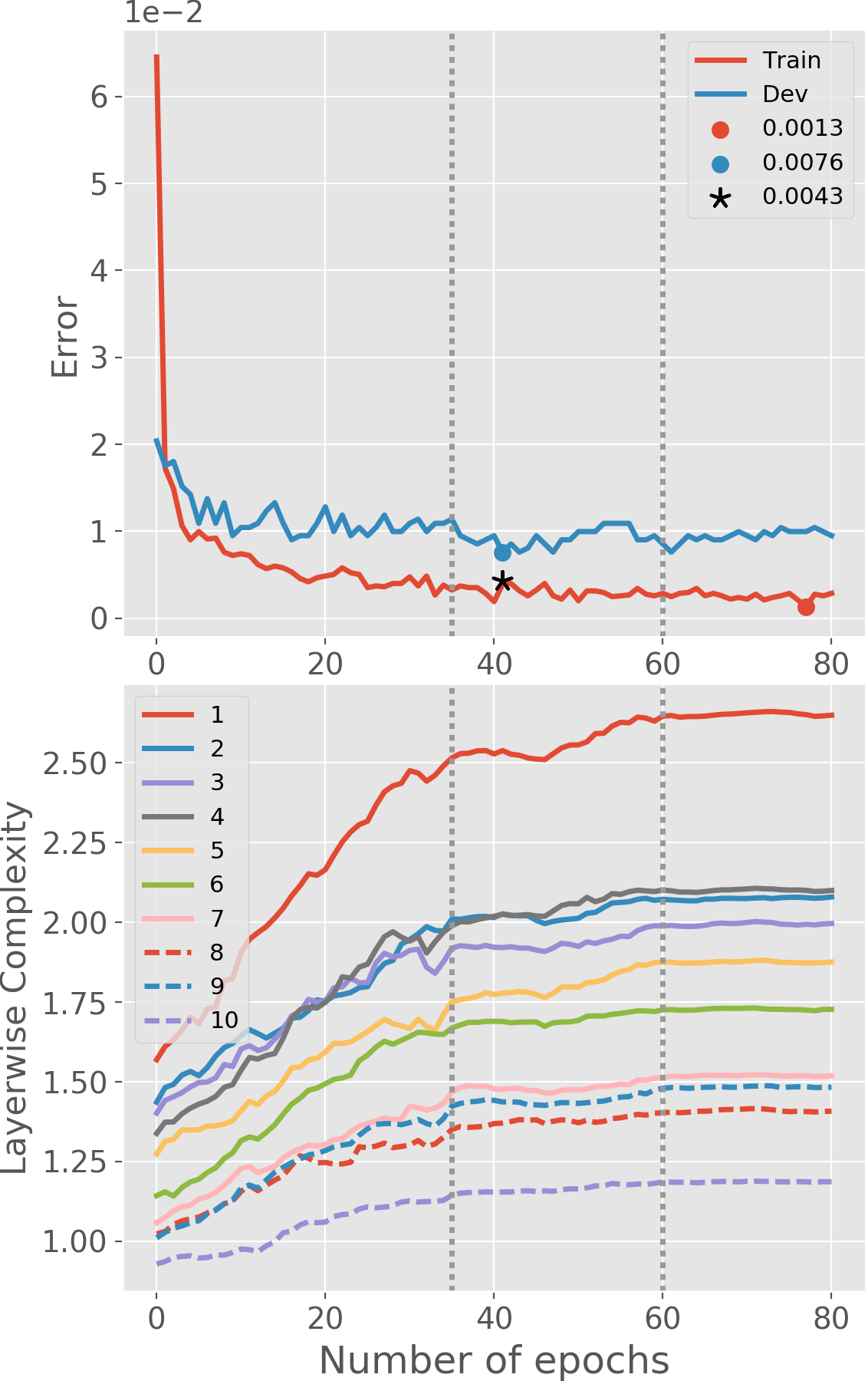}%
}
\subfloat[Ten Class]{
\includegraphics[width=0.49\columnwidth]{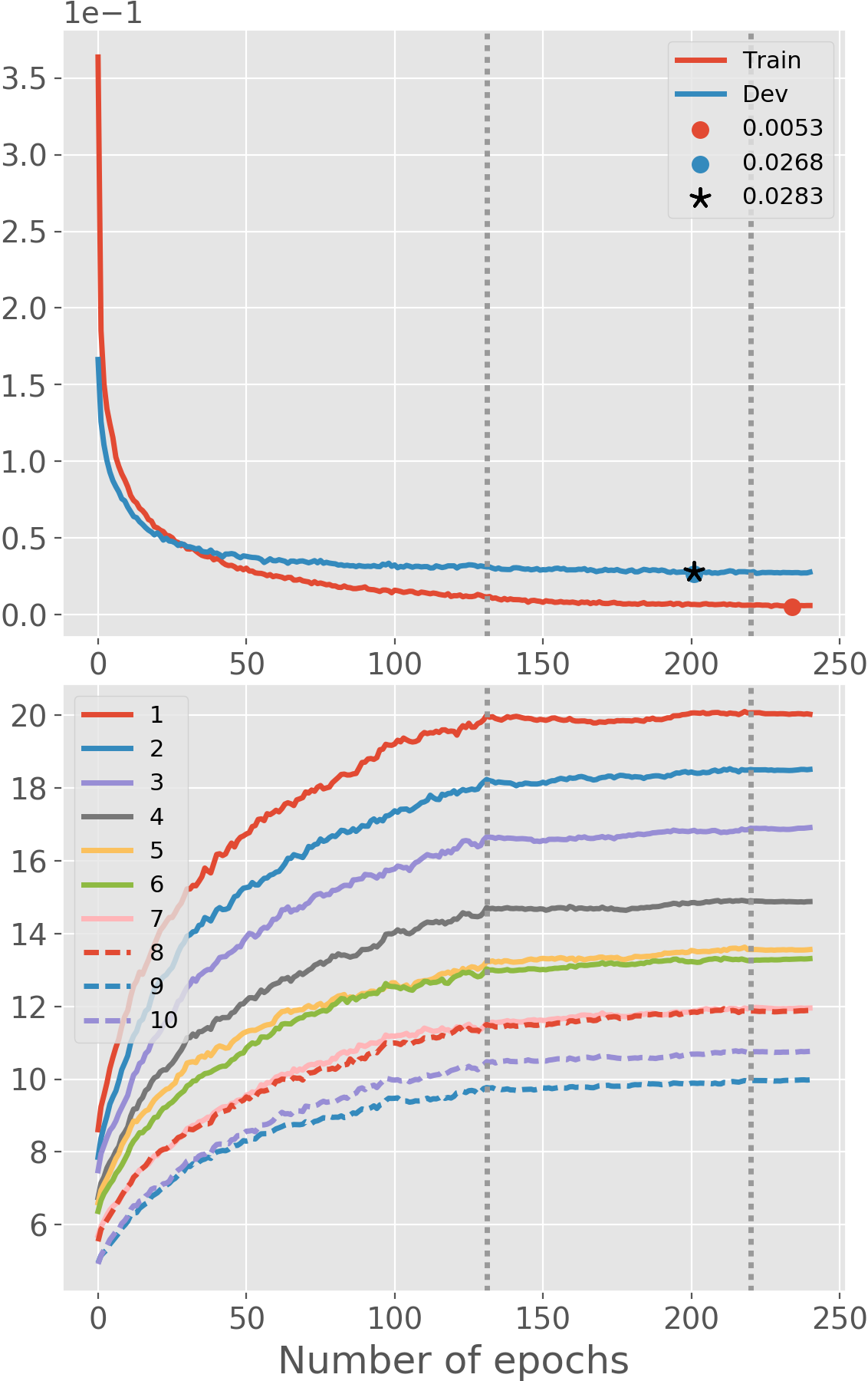}
}
\caption{Results on MNIST using highway networks.
Left and right columns correspond to the easy task (two classes) and the original task (ten classses).  Top and bottom rows show error rates and layerwise complexities for each layer. Training and validation errors (shortened as dev for development) are shown in red and blue. The best training and validation errors are marked with a dot. The test set error rate of the best performing model (on the validation set) is marked with an asterisk.}
\label{fig-mnist-hway}
\end{figure}

\begin{figure}[!htbp]
\centering
\subfloat[Two-class]{
\includegraphics[width=0.49\columnwidth]{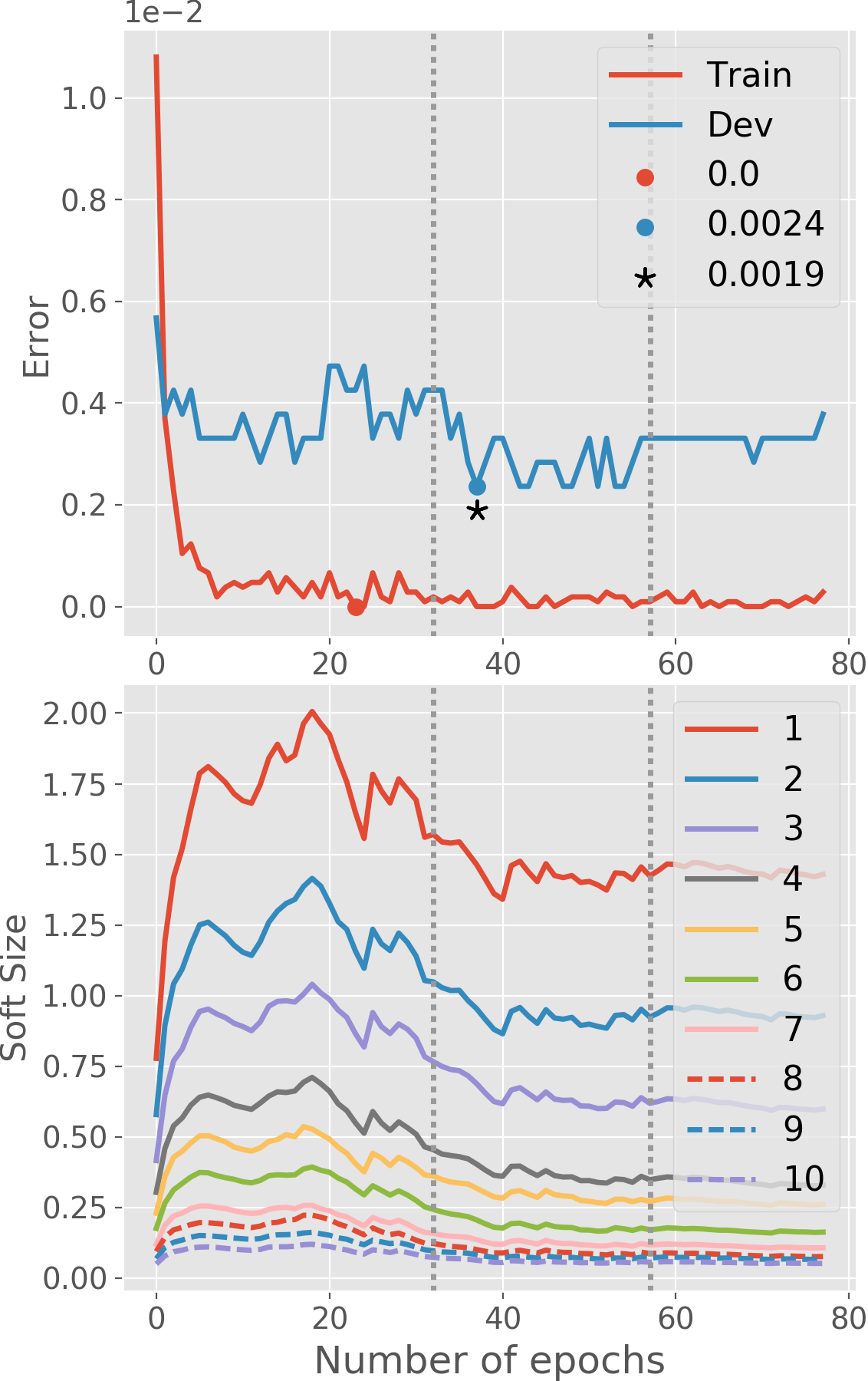}
}
\subfloat[Ten Class]{
\includegraphics[width=0.49\columnwidth]{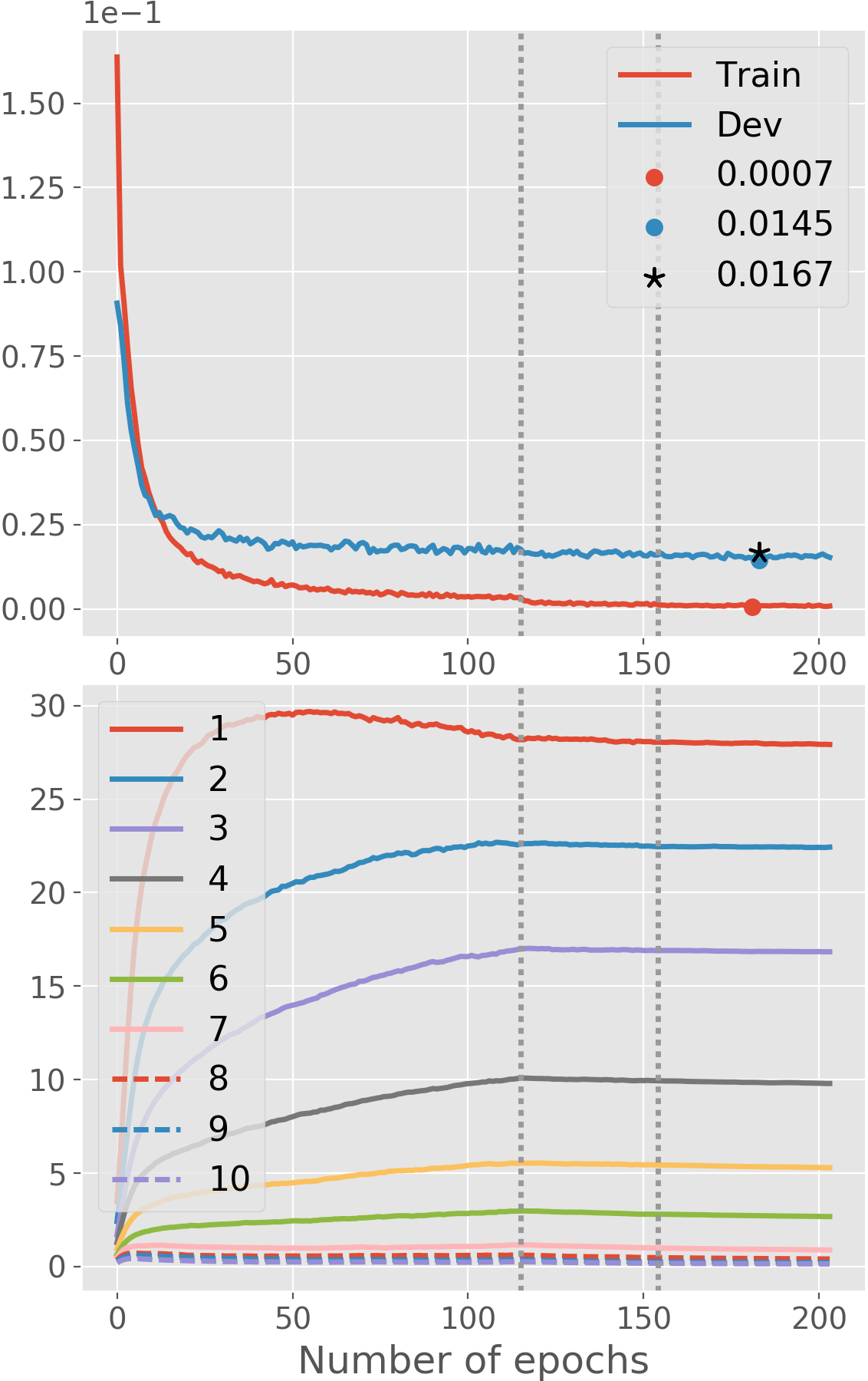}
}
\caption{Results on MNIST using tunnel networks.
Top and bottom rows show error rates and layerwise complexities for each layer.}
\label{fig-mnist-budmlp}
\end{figure}

\begin{figure}[!htbp]
\centering
\subfloat[Two-class]{
\includegraphics[width=0.48\columnwidth]{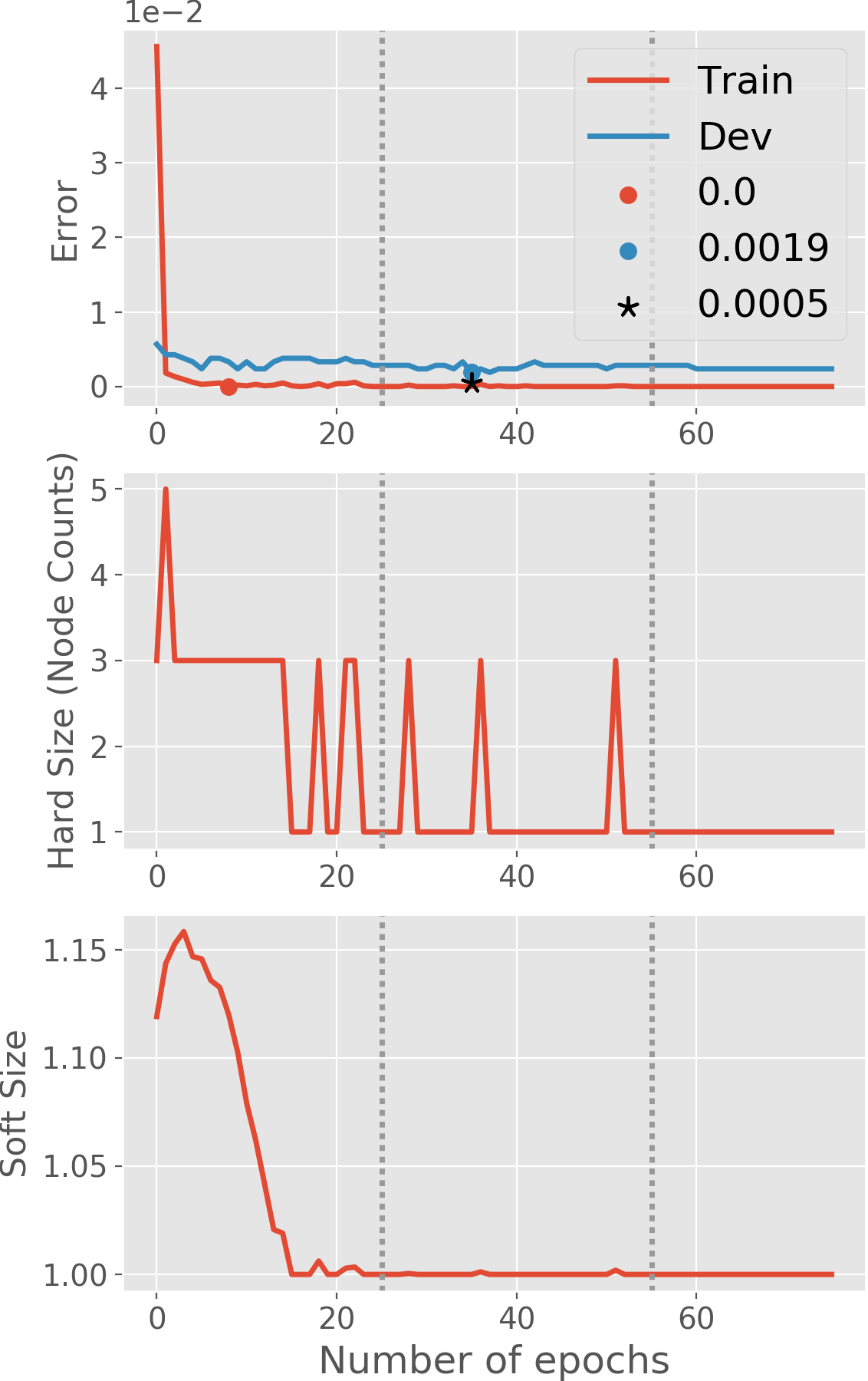}
}
\subfloat[Ten Class]{
\includegraphics[width=0.48\columnwidth]{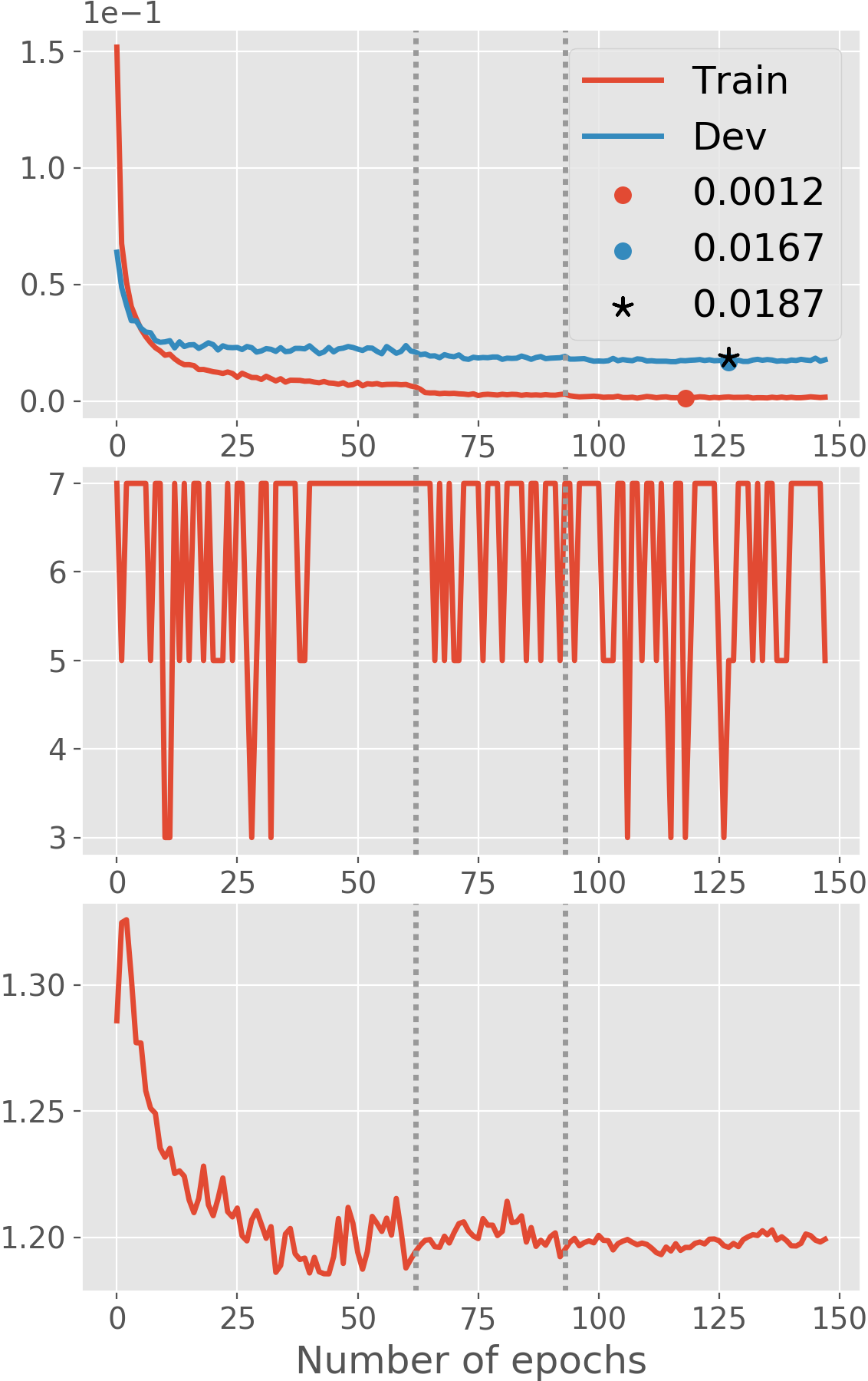}
}
\caption{Results on MNIST using budding perceptrons. Rows correspond to error rates, hard tree size and soft tree size, respectively.}
\label{fig-mnist-btn}
\end{figure}

\subsection{Results on Digit Recognition}

\textbf{Data.}
Next, we evaluate our constructive models on MNIST which contains 60,000 training and 10,000 test examples of handwritten digit images, each of which is $28\times 28$ pixels (784-dimensional) and belong to one of 10 classes~\cite{lecun1998mnist}. We randomly partition the original training set into training and validation sets with a ratio of 5:1. We also define the easier task of binary classification of separating digits 0 from 1, discarding all other classes from the data.

\textbf{Training.} We use the standard multi-class cross-entropy with a softmax output layer. We use 100 hidden units for each layer. Learning rate and L1 regularizer coefficient are fixed to 0.0003 and 0.001, respectively. An additional dropout regularizer is used at the input layer with a probability of 0.25 \cite{hinton2012}. All other hyperparameters are the same as before.

\textbf{Results.}
Our plots using highway networks, tunnel networks and budding perceptrons are shown in Figures \ref{fig-mnist-hway}, \ref{fig-mnist-budmlp}, and \ref{fig-mnist-btn}, respectively. We perform early stopping, i.e. the best model is chosen over all iterations based on the validation error rate and its test performance is evaluated. Training and validation errors are shown in red and blue, respectively. Best training and validation errors are marked with a dot. The test set error rate of the best performing model is marked with an asterisk. Note that for budding perceptrons our size measure is at the scale of depth (soft count of number of layers) whereas for tunnel networks and highway networks at the scale of hidden units (soft count of number of hidden units for each layer).

The top row of Figure~\ref{fig-mnist-hway} shows the training and validation errors with highway networks. On the easy 0 vs 1 task, it is possible to reach almost zero test error---the highway network gets 0.43\% (top left). On the ten-class task, training error is also very close to zero but it reaches a test error of 2.83\% suggesting overfitting. The bottom row shows the soft sizes of highway networks for individual layers. On both the easy task (left) and the ten-class task (right), we see that the model grows as it trains more until the learning rate shrinkage happens (shown with vertical dashed lines). However on neither of the tasks do we observe any pruning back of the model. Since there is no regularization term, highway networks do not have any incentive to reduce the model size.

Figure~\ref{fig-mnist-budmlp} shows similar statistics for the tunnel networks. The top row shows the training, validation and test errors on the two-class (left) and ten-class (right) tasks. The optimal performance is reached in about 40 epochs on the easy case whereas for the difficult case it takes about 175 epochs. On the ten-class task, the tunnel network can reach as low as 1.67\% error rate on the test set.

In the bottom row, we see layerwise soft sizes for the tunnel network. On the two-class task, layers grow and shrink together very consistently (curves seem to be mere translations/rescalings of one another). This is likely because the two-class case is very close to being, if not exactly, linearly separable. Hence, there is no need for the network to divide functionality among different layers differently. We also see that the maximum size ever reached is 2. On the other hand, on the ten-class case (right), the behavior is different and the peak size can reach 30 for the first layer. In both cases, earlier layers are used more actively.

Figure~\ref{fig-mnist-btn} shows results using budding perceptrons. As before, the top row shows error rates: The budding perceptron can reach near zero test error on the two-class task (left) and 1.84\% error rate on the ten-class task (right). Again, the optimum is reached much more quickly for the easy task, in about 40 epochs compared to 125. In terms of hard sizes, which are shown in the middle row, the budding perceptron seems to grow initially and then shrinks to the minimum size of 1 for the two-class task (left), whereas it reaches bigger sizes and stays almost flat for the ten-class task (right).

In the bottom row, we observe that both for the easy task (left) and the ten-class task (right), soft sizes grow initially and then the model is pruned back. Soft sizes for the ten class is slightly bigger: 1.2 vs 1. We also observe that the soft and hard sizes for the ten-class task do not look correlated, verifying the need to have the soft measure in addition to hard counts. Both hard and soft sizes reaching the minimum possible value of 1 for the two-class task provides another evidence for it to be very close to linearly separable.

We also observe that the learning rate is important for growth/shrinkage dynamics in all three models. When we shrink the learning rate (shown with vertical dashed bars), we see a stabilization over the sizes of the models even though performance continues to improve and after we shrink it a second time, sizes remain roughly constant.

\begin{figure*}[!htbp]
\centering
\includegraphics[width=0.98\textwidth]{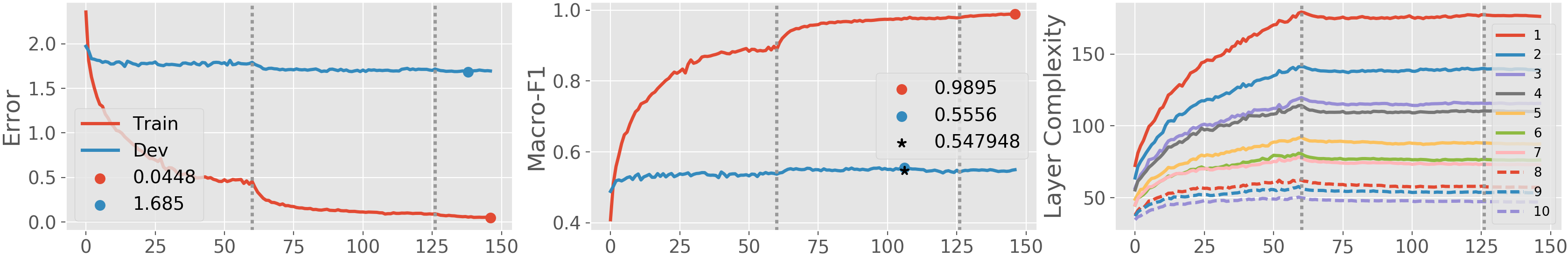}
\includegraphics[width=0.97\textwidth]{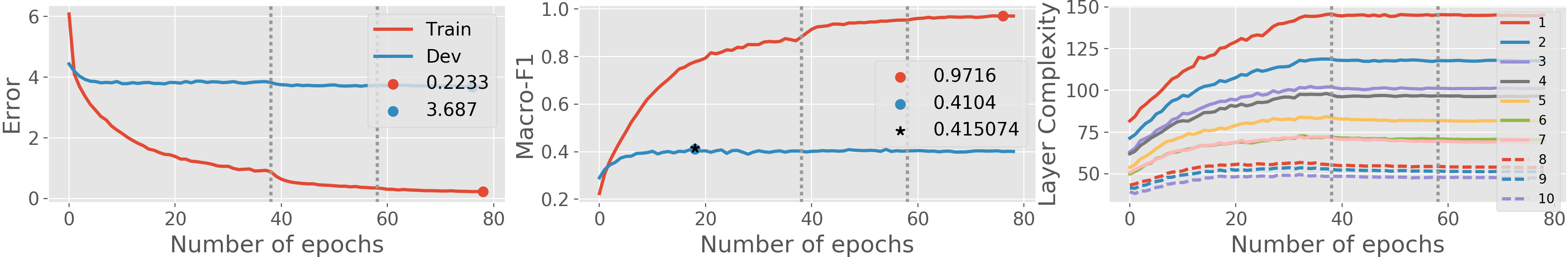}
\caption{Results on MIRFLICKR using highway networks. Top and bottom rows use easy task (10 labels) and original
task (38 labels), respectively.
Columns (left to right) correspond to error rates, Macro-F1, and layerwise complexities, respectively.}
\label{fig-flickr-hway}
\end{figure*}

\begin{figure*}[!htbp]
\centering
\includegraphics[width=0.97\textwidth]{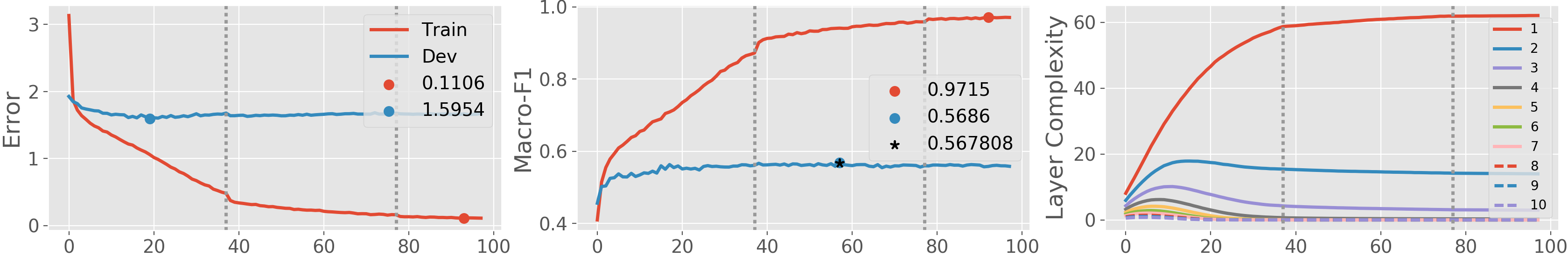}
\includegraphics[width=0.97\textwidth]{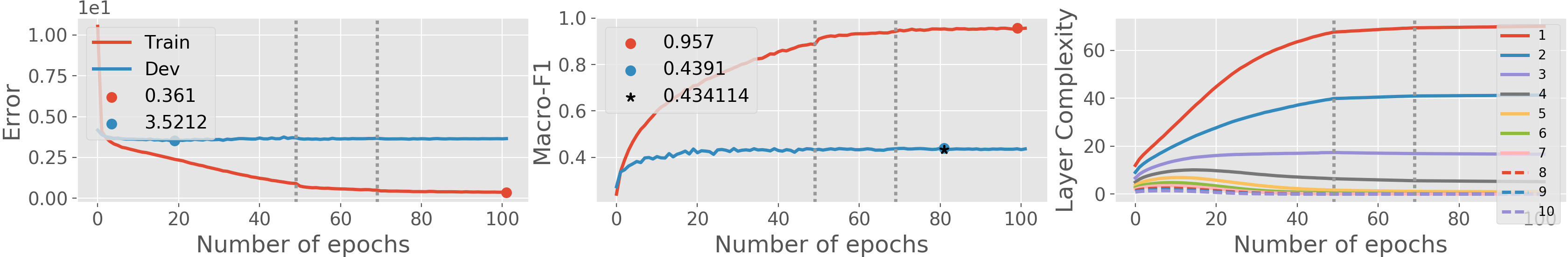}
\caption{Results on MIRFLICKR using tunnel networks. Top and bottom rows use easy task (10 labels) and original
task (38 labels), respectively.
Columns (left to right) correspond to error rates, Macro-F1, and layerwise complexities, respectively.}
\label{fig-flickr-budmlp}
\end{figure*}

\begin{figure*}[!htbp]
\centering
\includegraphics[width=0.97\textwidth]{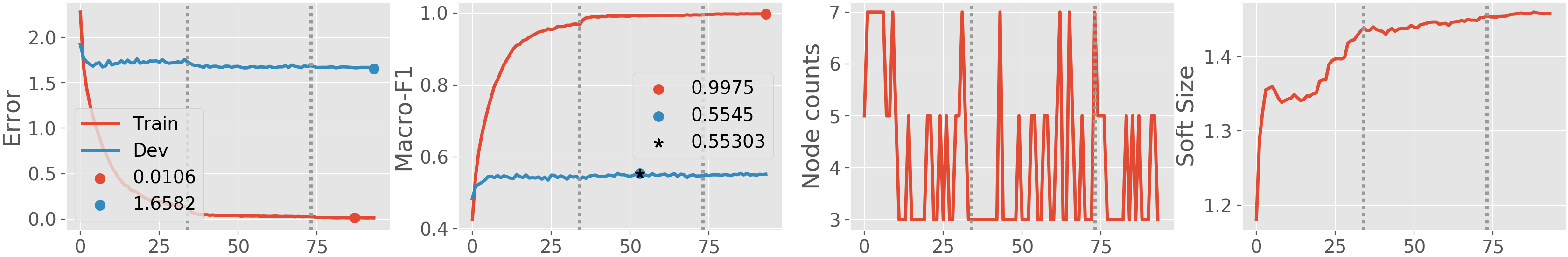}
\includegraphics[width=0.97\textwidth]{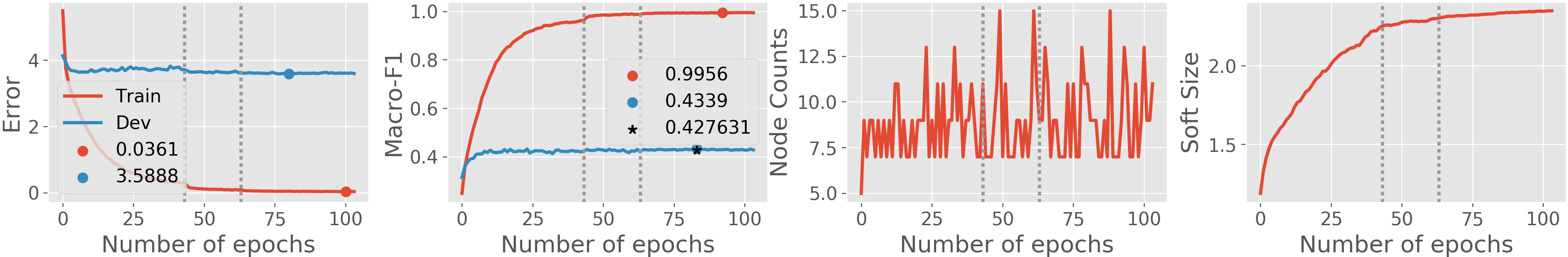}
\caption{Results on MIRFLICKR using budding perceptrons. Top and bottom rows use easy task (10 labels) and original
task (38 labels), respectively.
 Columns (left to right) correspond to error rates, Macro-F1, hard tree size and soft tree size, respectively.}
\label{fig-flickr-btn}
\end{figure*}

\subsection{Results on Image Tagging}

\textbf{Data.} We evaluate our models also on MIRFLICKR data set which has 25,000 Flickr raw images with associated topic labels~\cite{mirflickr}. We use the partitioning used in \cite{srivastava14b} with ratios 2:1:2 into training, validation and test sets. Output labels are the tags of images which are represented as a 38-dimensional vector of binary values (not necessarily one-hot); therefore the task is posed as a multi-label  classification problem where each instance might have multiple (or no) labels. Images are represented as a 3,857-dimensional feature vector, using the preprocessed and extracted image features as in~\cite{srivastava14b}.

To vary the difficulty of the data set and create an easier version of the problem, we use the label hierarchy to collapse the secondary labels~\cite{mirflickr}. For instance, all of the tags \emph{people}, \emph{baby}, \emph{female}, \emph{male}, and \emph{portrait} are collapsed into a single \emph{people} tag. This reduces the label space from 38 to 10.

\textbf{Training.} Learning rate scheduling is done similarly as in the previous experiments. For the objective function, we use a multi-label binary cross-entropy, since labels are not mutually exclusive. This is simply a sum over individual binary cross-entropy losses over each individual label. Our output layer, therefore, uses a multi-dimensional (elementwise) sigmoid function. We use a minibatch size of 32.

We use 300 hidden units in each layer. We fix the learning rate to 0.0003, which behaves well without oscillatory behavior for any of the models. For budding tree networks and tunnel networks, we use an L1 regularizer penalty of 0.1. We have an additional dropout regularization at the input layer with a probability of 0.25 \cite{hinton2012}.

\textbf{Performance measures.}
In addition to the total average error rate (summed over all labels) we use the macro F1-score, which is computed by averaging F1-scores for each individual label over all labels. Model selection is performed using Macro-F1 scores over the validation set.

\textbf{Results.}
Our plots for highway networks, tunnel networks and budding perceptrons are shown in Figures \ref{fig-flickr-hway}, \ref{fig-flickr-budmlp}, and \ref{fig-flickr-btn}, respectively. We show the Macro-F1 score on the test set (achieved by the best performing model on the validation set) with an asterisk.

In the top row of Figure~\ref{fig-flickr-hway}, we see the error rates (left), Macro-F1 scores (middle), and layer complexities (right) of highway networks on the easy task. We see that the model can reach a training error rate of 0.0448 per instance whereas the validation error rate is 1.685 (note that one instance can have more than one error in the multilabel case). We also see that the validation curve stays mostly flat during training. This suggests that the initial features might be too expressive making it easy to overfit. We see a similar phenomenon for the Macro-F1 scores (top middle), where training Macro-F1 score is 98.95\% whereas it is 55.56\% on the validation set.

On the difficult task, the behavior is very similar (bottom left and bottom middle): The error rates are 0.2233 for training vs. 3.687 for validation set, and Macro-F1 scores of 97.16\% for training and 41.04\% for validation set. The model sizes also show similar behavior (top right and bottom right) in terms of how each layer monotonically grows. Counterintuitively, the easy task reaches higher complexity values (for instance, about 175 for layer 1 for the easy task vs around 150 for the difficult task) which might be due to the aggressive early learning dynamics of the highway network.

We see similar statistics for tunnel networks in Figure~\ref{fig-flickr-budmlp}. In terms of error rates (left column) and Macro-F1 scores (middle column), a large gap between training and validation scores remain. Tunnel networks reach 97.15\% and 56.86\% training and validation Macro-F1 scores on the easy task, and 95.7\% and 43.91\% on the difficult task. For layer complexities (right column), tunnel networks show a different behavior than highway networks: Instead of using many layers the tunnel network utilizes mostly only the first two on the easy task (top right), and the first four on the difficult task (bottom right). We see that other layers see a small bump initially which is pruned afterwards. We also see that individual layer sizes (and hence the total size) is bigger on the difficult task.

Results using budding perceptrons are shown in Figure~\ref{fig-flickr-btn}. For error rates (first column) and Macro-F1 scores (second column), there is still a large difference between training and validation scores as with the previous two models. Budding perceptrons reach 99.75\% and 55.45\% training and validation Macro-F1 scores on the easy task, and 99.56\% and 43.39\% on the difficult task. In terms of the number of nodes (third column), easy task seems to consistently require between 7 and 3 nodes (full layers), mostly spending time on 3 and 5 (top). On the hard task however, we both see bigger values (always above or equal to 7 after the very first epoch) as well as a tendency for growth. For soft sizes (last column), the easy task shows an initial quick growth and then a pruning stage, which is followed by a more stable growth curve (top). On the hard task, we see a more consistent growth that is slowed by learning rate shrinkage (bottom).

\section{Conclusions and Future Work}
\label{sec:conclusions}

We propose two methods for learning the structure of a deep neural network where network complexity at the level of hidden unit or layer is coded by continuous parameters. These parameters are adjusted together with the network weights during gradient-descent, which implies softly modifying the network structure together with the network weights. Our contribution in this work is two-fold:
\begin{itemize}
    \item We propose tunnel networks as a simplified version of highway networks, and interpret unitwise $g$ parameters as a soft-notion of adding (or pruning) a unit to the network. We show that tunnel networks perform as effectively as highway networks while having much fewer parameters. 
    \item We propose budding perceptrons, which utilize a tree structure to represent a continuous and complete space of feedforward networks of arbitrary depth and optimize over this space using gradient descent. To our knowledge, this is the first constructive method that operates continuously and fully jointly, rather than making individual decisions to add or prune layers in stages.
\end{itemize}

Our experiments on the synthetic two-spirals data illustrate how tunnel networks and budding perceptrons can adapt to different sizes for different complexities of tasks using the same set of hyperparameters, by adapting the number of units for tunnel networks and the number of layers for budding perceptrons.

On the real-world tasks of digit recognition and image tagging, we have observed that tunnel networks achieve better performance by providing a better regularized model and using fewer number of parameters, compared to highway networks. We also observe that in all three tasks tunnel networks start by a growing exploratory phase, then shift into a pruning phase where the size is reduced.

Similarly, budding perceptrons have shown comparable or better performance on all three tasks. Compared to tunnel networks, budding perceptrons seem to grow more and prune less. A property of budding perceptron is its ability to grow to arbitrary depths instead of having to specify a maximum size. For instance, throughout this work we have used a maximum of 10 layers for highway and tunnel networks, however on MIRFLICKR, the budding perceptron sometimes reaches to a node count of 15.

By setting the learning rate in a decreasing manner, we ensure that different layers grow at different paces and are utilized differently. Combined with regularization, this allows for tunnel networks to keep some unused upper layers linear, allowing us the possibility to actually prune them from the network at the end. 

For future work, it will be interesting to see different application areas of our constructive neural networks. One potential approach is to incorporate convolutional layers within the constructive architecture, as typically used in computer vision applications~\cite{krizhevsky2012imagenet}, where the structure of the convolutional layers are also incrementally learned. Another possible direction is to apply the methods presented here in a sequential prediction setting, similar to recurrent highway networks \cite{zilly2016recurrent}; in such a case, a constructive model can learn how strongly and how much in the past previous information needs to be taken into account.

\ifCLASSOPTIONcaptionsoff
  \newpage
\fi

% trigger a \newpage just before the given reference
% number - used to balance the columns on the last page
% adjust value as needed - may need to be readjusted if
% the document is modified later
%\IEEEtriggeratref{8}
% The "triggered" command can be changed if desired:
%\IEEEtriggercmd{\enlargethispage{-5in}}

% references section

% can use a bibliography generated by BibTeX as a .bbl file
% BibTeX documentation can be easily obtained at:
% http://mirror.ctan.org/biblio/bibtex/contrib/doc/
% The IEEEtran BibTeX style support page is at:
% http://www.michaelshell.org/tex/ieeetran/bibtex/
%\bibliographystyle{IEEEtran}
% argument is your BibTeX string definitions and bibliography database(s)
%\bibliography{IEEEabrv,../bib/paper}
%
% <OR> manually copy in the resultant .bbl file
% set second argument of \begin to the number of references
% (used to reserve space for the reference number labels box)
\bibliographystyle{IEEEtran}
\bibliography{IEEEabrv,ref}

% biography section
% 
% If you have an EPS/PDF photo (graphicx package needed) extra braces are
% needed around the contents of the optional argument to biography to prevent
% the LaTeX parser from getting confused when it sees the complicated
% \includegraphics command within an optional argument. (You could create
% your own custom macro containing the \includegraphics command to make things
% simpler here.)
%\begin{IEEEbiography}[{\includegraphics[width=1in,height=1.25in,clip,keepaspectratio]{mshell}}]{Michael Shell}
% or if you just want to reserve a space for a photo:

%\begin{IEEEbiography}{Ozan \.Irsoy}
%Biography text here.
%\end{IEEEbiography}

%\begin{IEEEbiography}{Ethem Alpayd{\i}n}
%Biography text here.
%\end{IEEEbiography}

% insert where needed to balance the two columns on the last page with
% biographies
%\newpage

\end{document}